# One-Shot Coresets: The Case of $k$-Clustering

**Olivier Bachem**
ETH Zurich

**Mario Lucic**
Google Brain

**Silvio Lattanzi**
Google Research

## Abstract

Scaling clustering algorithms to massive data sets is a challenging task. Recently, several successful approaches based on data summarization methods, such as coresets and sketches, were proposed. While these techniques provide provably good and small summaries, they are inherently problem dependent — the practitioner has to commit to a fixed clustering objective before even exploring the data. However, can one construct small data summaries for a wide range of clustering problems *simultaneously*? We affirmatively answer this question by proposing an efficient algorithm that constructs such *one-shot* summaries for a large family of $k$-clustering problems while retaining strong theoretical guarantees.

## 1 Introduction

Clustering is a fundamental unsupervised learning task with a myriad of applications ranging from statistical data analysis and pattern recognition to data compression. The task is to partition a set of elements into groups such as to minimize some objective. The most popular algorithms, such as $k$-Means, compute clusterings which aim to minimize the expected (squared Euclidean) distance of a point to the closest cluster center. While computing the optimal clustering is usually NP-hard, a wide variety of provably good and practical approximation algorithms have been developed [2, 4, 5, 9].

While these algorithms perform well on sufficiently small data sets, they fail to scale to the massive data setting. To address this issue, several data summarization approaches were recently explored. The idea is very intuitive — instead of solving the problem on the full data set, one first summarizes the data set and then solves the problem on the summary. One of the most prominent approaches is based on *coresets* — small summaries of the data set which guarantee that the solution on the summary is competitive with the solution on the full data set [8]. Interestingly, for a fixed clustering problem, such as $k$-Means, one can efficiently compute a small coreset whose size is sublinear in, or even independent of the size of the data set [7, 17].

The main drawback of these approaches is that the coreset construction is inherently tied to the clustering problem. This is a significant drawback, as one usually explores several objective functions and chooses the one which best fits the application at hand. For example, given a data set drawn from a mixture of spherical Gaussians, $k$-Means is a well-suited objective. However, if one adds uniform random noise to this data set, one might want to consider $k$-Medians to ensure robustness. In general, selecting the best clustering objective a priori is hard, and one usually considers several clustering objectives. Hence, a natural question is whether one can construct a *small* coreset that is provably good for all $k$-clustering problems simultaneously?

**Our contributions.** In this work, we

1. introduce and motivate the concept of *one-shot coresets*, i.e., data summaries suitable for multiple clustering problems simultaneously,

2. propose and theoretically analyze an efficient algorithm to construct such summaries for a large family of $k$-clustering problems, and

3. improve on state of the art coreset constructions for both metric and Euclidean spaces.

## 2  Background and Related Work

**$k$-clustering problem.** Let $\mathcal{X}$ denote a set of $n$ equally weighted points and let $\mathrm{d} : \mathcal{X} \times \mathcal{X} \to [0, \infty)$ be a metric defined on $\mathcal{X}$. The goal of *metric $k$-clustering* with $p \in [1, \infty)$ is to select a set $Q \subseteq \mathcal{X}$ of at most $k$ cluster centers which minimizes

$$\phi_{\mathcal{X}}^{p}(Q) = \frac{1}{|\mathcal{X}|} \sum_{x \in \mathcal{X}} \mathrm{d}(x, Q)^{p} = \frac{1}{|\mathcal{X}|} \sum_{x \in \mathcal{X}} \min_{q \in Q} \mathrm{d}(x, q)^{p}.$$

Similarly, for a weighted set $\mathcal{C}$ with weight function $w : \mathcal{C} \to \mathbb{R}_{\geq 0}$, we define $\phi_{\mathcal{C}}^{p}(Q) = \sum_{x \in \mathcal{C}} w(x)\,\mathrm{d}(x, Q)^{p}$. The *Euclidean $k$-clustering* problem is to find a set of at most $k$ centers in $\mathbb{R}^{d}$ that minimizes

$$\phi_{\mathcal{X}}^{p}(Q) = \frac{1}{|\mathcal{X}|} \sum_{x \in \mathcal{X}} \min_{q \in Q} \|x - q\|_{2}^{p}.$$

This formalization generalizes the most popular clustering problems such as $k$-Median (p=1), $k$-Means (p=2) and $k$-Center ($p \to \infty$). We note that in general metric spaces the cluster centers have to be data points, whereas in Euclidean spaces they can be arbitrary points in $\mathbb{R}^{d}$.

**$D^{p}$-sampling.** In general, the $k$-clustering problem is NP-hard, even in the Euclidean plane with $p = 2$ [13]. However, it is possible to obtain an approximation of the optimal solution using $D^{p}$-sampling [2] as described in Algorithm 1: One first samples an initial cluster center uniformly at random from the data set $\mathcal{X}$. Then, in each of $k - 1$ subsequent iterations, a data point is sampled as a new cluster center with probability proportional to $\mathrm{d}(\cdot, C)^{p}$, where $C$ is the set of already sampled cluster centers. For any fixed $p$, Algorithm 1 returns a solution which is $\mathcal{O}(2^{p} \log k)$ competitive with the optimal solution in expectation [2].

**Coresets.** Coresets are a proven way to scale several instances of the $k$-clustering problem [10, 11, 14, 17]. The idea is to compute a small representative summary $\mathcal{C}$ of the original data set $\mathcal{X}$, such that the underlying objective function, e.g. $\phi_{\mathcal{X}}^{p}(Q)$, is well approximated by the objective based on the summary $\mathcal{C}$, i.e., $\phi_{\mathcal{C}}^{p}(Q)$. One may then show that solutions obtained on the summary $\mathcal{C}$ are provably competitive with solutions trained on $\mathcal{X}$ [8]. For many problems, such as $k$-Median or $k$-Means, the coreset size is *independent* of the size of $\mathcal{X}$, which makes them a great candidate for large-scale inference as more computationally expensive methods may be used on the small summary [8]. Furthermore, coresets satisfy strong compositional properties [1, 14]. As soon as one has a valid coreset construction, one essentially obtains efficient streaming and parallel coreset constructions for free.

A key drawback of the state-of-the-art coreset constructions is that the clustering problem needs to be specified *a priori*. From the practitioners point of view, this is a drawback, as one might like to explore several objective functions and then choose the one which best fits the application at hand. While one might try to construct different coresets for different clustering problems, there are substantial drawbacks: One might want to explore an unbounded number of clustering problems simultaneously, making a brute-force approach unsuitable. Similarly, large data sets can often only be stored in distributed systems or arrive as a stream. As such, it is advantageous to directly construct a single coreset that provides guarantees for multiple objective functions simultaneously, instead of sequentially constructing coresets for specific clustering problems as the data is explored.

---

**Algorithm 1** $D^{p}$-Sampling

**Require:** data set $\mathcal{X}$, $k$, $p$
1: Uniformly sample $x \in \mathcal{X}$ and set $B = \{x\}$.
2: **for** $i \leftarrow 2, 3, \ldots, k$ **do**
3:     Sample $x \in \mathcal{X}$ with probability
       $\frac{\mathrm{d}(x, B)^{p}}{\sum_{x' \in \mathcal{X}} \mathrm{d}(x', B)^{p}}$ and add it to $B$.
4: **return** $B$

---

## 3  Coresets for $k$-Clustering in Metric and Euclidean Spaces

We first consider the $k$-clustering problem for a fixed value for $p$. We introduce a coreset construction algorithm that results in state-of-the-art coreset sizes for both Euclidean $k$-clustering and general metric $k$-clustering. This algorithm will serve as the foundation for the proposed *one-shot coresets* in Section 4.

There is a well-established notion of coresets for the $k$-clustering problems $k$-Median and $k$-Means [3, 10, 17, 18]. Intuitively, the cost function $\phi_{\mathcal{C}}^{p}(Q)$ has to approximate the true cost $\phi_{\mathcal{X}}^{p}(Q)$ up to a multiplicative factor of $1 \pm \varepsilon$. For Euclidean spaces, Definition 1 generalizes this notion to $k$-clustering with arbitrary, but constant values of $p$.

**Definition 1** (Coreset for Euclidean $k$-clustering). *Let $\varepsilon > 0$, $k \in \mathbb{N}$ and $p \in [1, \infty)$. Let $\mathcal{X} \subset \mathbb{R}^d$ be a set of $n$ points. The weighted set $\mathcal{C}$ is an $\varepsilon$-coreset of $\mathcal{X}$ for $k$-clustering in $\mathbb{R}^d$ with power $p$ if, for any set $Q \subset \mathbb{R}^d$ of cardinality at most $k$,*

$$|\phi_{\mathcal{X}}^p(Q) - \phi_{\mathcal{C}}^p(Q)| \leq \varepsilon \phi_{\mathcal{X}}^p(Q). \tag{1}$$

Similarly, an established notion of coresets for metric spaces given in [12] is generalized in Definition 2.

**Definition 2** (Coreset for metric $k$-clustering). *Let $\varepsilon > 0$, $k \in \mathbb{N}$ and $p \in [1, \infty)$. Let $\mathcal{X}$ be an arbitrary set of $n$ points and $\mathrm{d}(\cdot, \cdot)$ a metric on $\mathcal{X}$. The weighted set $\mathcal{C}$ is an $\varepsilon$-coreset of $\mathcal{X}$ for $k$-clustering with power $p$ if, for any set $Q \subseteq \mathcal{X}$ of cardinality at most $k$,*

$$|\phi_{\mathcal{X}}^p(Q) - \phi_{\mathcal{C}}^p(Q)| \leq \varepsilon \phi_{\mathcal{X}}^p(Q). \tag{2}$$

### 3.1 Coreset construction algorithm

We first provide the intuition behind the coreset construction and then present the algorithm in detail. The objective function $\phi_{\mathcal{X}}^p(Q)$ may be reformulated as the expected loss of a single point $x$ drawn uniformly at random from the data set $\mathcal{X}$. Hence, drawing a random subsample of $\mathcal{X}$ and evaluating a solution on that random subsample provides an unbiased estimator of the true objective function. However, since the loss of data points can be very different, random subsampling does not guarantee the existence of a small coresets as defined in Definitions 1 or 2 [8]. In fact, in the worst case, the subsample would have to be of size $\Theta(n)$, even for only a single query $Q$. In the presence of unbalanced clusters and heavy-tailed data the issue becomes ever more prominent.

The standard remedy to this difficulty is to use importance sampling where points with a potentially high impact on the objective function are sampled with a higher probability, but assigned a correspondingly lower weight [8, 15]. Formally, let the *sensitivity* of $x \in \mathcal{X}$ be defined as in [15], i.e.,

$$\sigma_p(x) := \sup_{Q \in \mathcal{Q}} \frac{d(x, Q)^p}{\frac{1}{|\mathcal{X}|} \sum_{x' \in \mathcal{X}} d(x', Q)^p}, \tag{3}$$

where $\mathcal{Q} = \mathbb{R}^{d \times k}$ for Euclidean spaces and $\mathcal{Q} = \mathcal{X}^k$ for general metric spaces. The *sensitivity* measures the worst-case ratio between the loss of $x$ and the average loss over all possible queries $Q$. While the

---

**Algorithm 2** SENSITIVITY

**Require:** data set $\mathcal{X}$, $p$, $\delta$, $k$
1: $B \leftarrow$ Best solution of $\ln \frac{1}{\delta}$ runs of Algorithm 1.
2: **for each** $b_i$ in $B$ **do**
3:    $B_i \leftarrow$ Set of points from $\mathcal{X}$ closest to $b_i$. Ties broken arbitrarily but consistently.
4: $\alpha \leftarrow 2^{p+3} (\log_2 k + 2)$
5: **for each** $b_i \in B$ and $x \in B_i$ **do**
6:    $s(x) \leftarrow \frac{\alpha 2^p d(x, b_i)^p}{2\phi_{\mathcal{X}}^p(B)} + \frac{\alpha 4^p \phi_{B_i}^p(b_i)}{4\phi_{\mathcal{X}}^p(B)} + \frac{4^p |\mathcal{X}|}{4|B_i|}$
7: **return** $s(\cdot)$

---

**Algorithm 3** CORESET FOR FIXED $p$

**Require:** data set $\mathcal{X}$, $p$, $\delta$, $m$, $k$
1: $s_p(\cdot) \leftarrow$ Run Algorithm 2 with $\delta/2$.
2: **for each** $x \in \mathcal{X}$ **do**
3:    $q(x) \leftarrow s_p(x) / \sum_{x' \in \mathcal{X}} s_p(x')$
4: $\mathcal{C} \leftarrow$ Sample $m$ weighted points from $\mathcal{X}$ where each point $x$ has weight $\frac{1}{|\mathcal{X}| m q(x)}$ and is sampled with probability $q(x)$.
5: **return** $\mathcal{C}$

---

sensitivity itself may be hard to compute exactly, it suffices to compute an *uniform upper bound* on the sensitivity [8, 15]. To this end, we propose Algorithm 2 which by the following Lemma *provably* computes such an upper bound $s(\cdot)$.

**Lemma 1.** *For $k \in \mathbb{N}$, $p \geq 1$, $\delta \in (0, 1)$ and a data set $\mathcal{X}$ of $n$ points, let $s : \mathcal{X} \to \mathbb{R}_{\geq 0}$ be the function returned by a single run of Algorithm 2. Then, with probability at least $1 - \delta$, $s_p(x) \geq \sigma_p(x)$, $\forall x \in \mathcal{X}$, and $S = \frac{1}{|\mathcal{X}|} \sum_{x \in \mathcal{X}} s_p(x) \leq 8^p k$. The computational complexity of Algorithm 2 is $\mathcal{O}(ndk \log \frac{1}{\delta})$.*

The proof is presented in the Appendix. Algorithm 2 effectively extends the sensitivity bound of [17] to both metric and Euclidean $k$-clustering. Intuitively, we first compute a rough approximation $B$ to the optimal clustering using $D^p$-sampling and then use that solution to bound the impact of each data point exploiting the triangle inequality in the underlying metric space.

The resulting coreset construction for $k$-clustering is detailed in Algorithm 3. We first compute the sensitivity bound $s(\cdot)$ using Algorithm 2 and then perform importance sampling proportional to $s(\cdot)$. The total computational complexity of Algorithm 2 is $\mathcal{O}(ndk \log \frac{1}{\delta})$. In Sections 3.2 and 3.3, we prove that Algorithm 3 computes small coresets for $k$-clustering in Euclidean and general metric spaces.

## 3.2 Coreset size for general metric spaces

**Theorem 1.** *Let $\varepsilon \in (0,1)$, $\delta \in (0,1)$ and $p \in [1,\infty)$. Let $\mathcal{X}$ be a data set of $n$ points and $\mathrm{d}(\cdot,\cdot)$ a metric on $\mathcal{X}$. Then, for*

$$m \geq \frac{8^{p+3}k}{3\varepsilon^2}\left(1 + k\log n + \log\frac{2}{\delta}\right)$$

*the weighted set $\mathcal{C}$ returned by Algorithm 3 is a $\varepsilon$-coreset of $\mathcal{X}$ for $k$-clustering with $\mathrm{d}(\cdot,\cdot)^p$, with probability at least $1 - \delta$.*

Theorem 1 implies that the coreset size is only logarithmic in the number of data points $n$ and in $\frac{1}{\delta}$ as well as quadratic in both $k$ and $\frac{1}{\varepsilon}$. The exponential dependence on $p$ quantifies the inherent difficulty of $k$-clustering problems for general $p$.

For both $k$-Median and $k$-Means, Theorem 1 implies a coreset size of $\mathcal{O}\left(\frac{k}{\varepsilon^2}\left(k\log n + \log\frac{1}{\delta}\right)\right)$. This improves the result in Chen [12] by a $\log n$ factor. Additionally, Algorithm 3 is more practical than the one in Chen [12], works provably well for any constant $p$, and has a substantially simpler proof.

*Proof of Theorem 1.* By Definition 2, we have to show that Algorithm 3, with probability at least $1 - \delta$, computes a weighted set $\mathcal{C}$ which satisfies

$$|\phi_\mathcal{X}^p(Q) - \phi_\mathcal{C}^p(Q)| \leq \varepsilon \phi_\mathcal{X}^p(Q), \quad (4)$$

or equivalently with $S = \frac{1}{|\mathcal{X}|}\sum_{x \in \mathcal{X}} s_p(x)$

$$\left|\frac{1}{S} - \frac{\phi_\mathcal{C}^p(Q)}{S\phi_\mathcal{X}^p(Q)}\right| \leq \frac{\varepsilon}{S}, \quad (5)$$

*uniformly* for all $Q \subset \mathcal{X}$ with $|Q| = k$. As Lemma 1 does not hold with probability at most $\delta/2$, our goal is to prove that (5) fails with probability at most $\delta/2$. Hence, for the remainder of the proof we may assume that Lemma 1 holds.

Consider the function

$$g_Q^p(x) = \frac{\mathrm{d}(x,Q)^p}{\phi_\mathcal{X}^p(Q)s_p(x)} \quad (6)$$

and note that, by Lemma 1 and the non-negativity of $\mathrm{d}(\cdot,\cdot)$, $g_Q^p(x)$ is bounded in $[0,1]$. Furthermore, we have that

$$\mathbb{E}\left[g_Q^p(x)\right] = \sum_{x \in \mathcal{X}} q(x)g_Q^p(x) = \frac{1}{S} \quad (7)$$

where the expectation is taken with respect to the distribution $q(x) = \frac{s_p(x)}{|\mathcal{X}|S}$. Furthermore, since each point $x \in \mathcal{C}$ has weight $w(x) = \frac{S}{|\mathcal{C}|s_p(x)}$, we have

$$\phi_\mathcal{C}^p(Q) = \sum_{x \in \mathcal{C}} w(x)\,\mathrm{d}(x,Q)^p = \frac{1}{|\mathcal{C}|}\sum_{x \in \mathcal{C}} \frac{S\,\mathrm{d}(x,Q)^p}{s_p(x)},$$

and thus

$$\frac{\phi_\mathcal{C}^p(Q)}{S\phi_\mathcal{X}^p(Q)} = \frac{1}{|\mathcal{C}|}\sum_{x \in \mathcal{C}} g_Q^p(x). \quad (8)$$

By (7) and (8), to prove (5) we have to show that

$$\left|\mathbb{E}\left[g_Q^p(x)\right] - \frac{1}{|\mathcal{C}|}\sum_{x \in \mathcal{C}} g_Q^p(x)\right| \leq \frac{\varepsilon}{S}, \quad (9)$$

*uniformly* for all $Q \subset \mathcal{X}$ with $|Q| = k$.

We first show that (9) holds for a single solution $Q$ and then extend it to all solutions $Q \subset \mathcal{X}$ with $|Q| = k$. Let $Y_1, Y_2, \ldots, Y_m$ be zero-mean random variables with $|Y_i| \leq 1$ for all $i$. Then, by the Bernstein inequality in [19], it holds for any $t > 0$ that

$$\mathbb{P}\left[\sum_{i=1}^m Y_i > t\right] \leq \exp\left(-\frac{\frac{1}{2}t^2}{\sum_{i=1}^m \mathbb{E}[Y_i^2] + \frac{t}{3}}\right). \quad (10)$$

For $i = 1, 2, \ldots, m$, let $x_i$ be the $i$-th sampled point in $\mathcal{C}$ and define the zero-mean random variables

$$Y_i = g_Q^p(x_i) - \frac{1}{S}. \quad (11)$$

Since both $g_Q^p(x_i)$ and $\frac{1}{S}$ are bounded in $[0,1]$, we have $|Y_i| \leq 1$ for all $i$ as well as

$$\sum_{i=1}^m \mathbb{E}\left[Y_i^2\right] \leq \sum_{i=1}^m \mathbb{E}\left[g_Q^p(x_i)^2\right] \leq \sum_{i=1}^m \mathbb{E}\left[g_Q^p(x_i)\right] = \frac{m}{S}. \quad (12)$$

Applying (10) with $t = \frac{m\varepsilon}{S}$ to both $Y_i$ and $-Y_i$ in combination with a union bound implies that

$$\left|\mathbb{E}\left[g_Q^p(x)\right] - \frac{1}{|\mathcal{C}|}\sum_{x \in \mathcal{C}} g_Q^p(x)\right| > \frac{\varepsilon}{S} \quad (13)$$

with probability at most

$$2\exp\left(-\frac{\frac{1}{2}\frac{m^2\varepsilon^2}{S^2}}{\frac{m}{S} + \frac{m}{S}\frac{\varepsilon}{3}}\right) \stackrel{(\varepsilon \leq 1)}{\leq} 2\exp\left(-\frac{3}{8}\frac{m\varepsilon^2}{S}\right) \quad (14)$$

for any single $Q$.

By the union bound, the probability that there exists a $Q \subseteq \mathcal{X}$ with $|Q| \leq k$ such that (13) holds is hence bounded by

$$2n^k \exp\left(-\frac{3}{8}\frac{m\varepsilon^2}{S}\right) \qquad (15)$$

since there are at most $n^k$ subsets of $\mathcal{X}$ of size at most $k$. By Lemma 1, we further have that $S \leq 8^{p+2}k$. Substituting $m$ and $S$ into (15) and simple arithmetic manipulations yield that (13) holds with probability at most $\delta/2$. This implies that (9) holds with probability at least $1 - \delta$ *uniformly* for all $Q \subseteq \mathcal{X}$ with $|Q| \leq k$ which concludes the proof. $\square$

### 3.3 Analysis for Euclidean spaces

**Theorem 2.** *Let $\varepsilon \in (0,1)$, $\delta \in (0,1)$ and $p \in [1, \infty)$. Let $\mathcal{X}$ be a data set of $n$ points in $\mathbb{R}^d$ and choose*

$$m \geq c\frac{8^p pk \log k}{\varepsilon^2}\left(dk \log k + \log \frac{1}{\delta}\right),$$

*where $c > 0$ is an absolute constant. Then, the weighted set $\mathcal{C}$ returned by Algorithm 3 is a $\varepsilon$-coreset of $\mathcal{X}$ for $k$-clustering in $\mathbb{R}^d$ with probability at least $1 - \delta$.*

In contrast to metric spaces, the coreset size in Euclidean spaces is independent of the number of data points $n$. For the case of $k$-Means clustering the coreset size is of $\mathcal{O}\left(\frac{k}{\varepsilon^2}\left(dk \log^2 k + \log \frac{1}{\delta}\right)\right)$ which improves the result of [17] by replacing a factor $k$ with $\log k$. The proof of Theorem 2 is provided in the Appendix and uses a generalization of the Vapnik-Chervonenkis dimension to $[0,1]$-valued functions to avoid a dependence on the number of data points $n$ as in Theorem 1.

## 4 One-shot Coresets for $k$-clustering

In this section, we investigate the question whether one can compute small coresets for a large, infinite family of $k$-clustering problems *simultaneously*. A straightforward approach would be to simply compute a single coreset for a fixed $p$ and then use that coreset for all desired values of $p$. However, as $p$ varies, the importance of points might wildly change. For example, for $k$-Median ($p = 1$) few far away points may safely be discarded while it is essential to include them for $k$-clustering with larger values of $p$. As such, it is critical to obtain a set $\mathcal{C}$ that provides theoretical guarantees for different values of $p$. Our main contribution is to show that it is possible to compute such coresets for a large, infinite family of $k$-clustering problems *simultaneously*. Analogous to the Definitions 1 and 2, we propose the following intuitive notion of *one-shot coresets* for $k$-clustering with *all $p \in [1, p_{max}]$* where the practitioner may choose $p_{max} \in [1, \infty)$.

**Definition 3** (Metric one-shot coresets). *Let $\varepsilon > 0$, $k \in \mathbb{N}$ and $p_{max} \in [1, \infty)$. Let $\mathcal{X}$ be an arbitrary set of $n$ points and $d(\cdot, \cdot)$ a metric on $\mathcal{X}$. The weighted set $\mathcal{C}$ is an $\varepsilon$-coreset of $\mathcal{X}$ for $k$-clustering with $d(\cdot, \cdot)^p$ and power $p \in [1, p_{max}]$ if for any set $Q \subseteq \mathcal{X}$ of cardinality at most $k$ and for any $p \in [1, p_{max}]$*

$$|\phi_{\mathcal{X}}^p(Q) - \phi_{\mathcal{C}}^p(Q)| \leq \varepsilon \phi_{\mathcal{X}}^p(Q). \qquad (16)$$

This novel notion of coresets can be similarly extended to Euclidean spaces.

**Definition 4** (Euclidean one-shot coresets). *Let $\varepsilon > 0$, $k \in \mathbb{N}$ and $p_{max} \in [1, \infty)$. Let $\mathcal{X} \subset \mathbb{R}^d$ be a set of $n$ points. The weighted set $\mathcal{C}$ is an $\varepsilon$-coreset of $\mathcal{X}$ for $k$-clustering in Euclidean spaces and power $p \in [1, p_{max}]$ if for any set $Q \subset \mathbb{R}^d$ of cardinality at most $k$ and for any $p \in [1, p_{max}]$*

$$|\phi_{\mathcal{X}}^p(Q) - \phi_{\mathcal{C}}^p(Q)| \leq \varepsilon \phi_{\mathcal{X}}^p(Q). \qquad (17)$$

The key difficulty is that (16) and (17) have to hold for all $p \in [1, p_{max}]$ simultaneously.

### 4.1 Coreset construction algorithm

We propose Algorithm 4 which computes *one-shot coresets* for $k$-clustering in both Euclidean and general metric spaces. It works as follows: We first cover the interval $[1, p_{max}]$ with the exponential grid $P = \{1, (1+\Delta), (1+\Delta)^2, \ldots, p_{max}\}$ where $p_{max}$ is chosen by the practitioner. Then, we use Algorithm 2 to compute an upper bound $s_p(x)$ on the sensitivity for each $p$ in the grid $P$. Finally, we construct a coreset by performing importance sampling proportional to $s(x) = \sum_{p \in P} s_p(x)$.

We first show that the choice of $\Delta = \frac{1}{\log n}$ is sufficient to guarantee that Algorithm 4 computes valid coresets for general metric spaces.

**Theorem 3** (Metric one-shot coresets). *Let $k \in \mathbb{N}$, $\varepsilon \in (0,1)$, $\delta \in (0,1)$ and $p \in [1, p_{max}]$. Let $\mathcal{X}$ be a data set of $n$ points and $d(\cdot, \cdot)$ a metric on $\mathcal{X}$.*

Then, for $\Delta = \frac{1}{\log n}$ and

$$m \geq c\frac{16^{p_{max}}k\log n}{\varepsilon^2}\left(k\log n + \log\frac{1}{\delta\varepsilon}\right),$$

where $c > 0$ is an absolute constant, the weighted set $\mathcal{C}$ returned by Algorithm 4 is a $\varepsilon$-coreset of $\mathcal{X}$ for $k$-clustering for all $p \in [1, p_{max}]$ with $d(\cdot,\cdot)^p$, with probability at least $1 - \delta$. The computational complexity of Algorithm 4 is $\mathcal{O}(nkd\log n\log\frac{1}{\delta}\log p_{max})$.

Compared to Theorem 1, we require an additional $\log n$ and $\log \frac{1}{\varepsilon}$ factor in the coreset size. Critically, however, Theorem 3 still leads to small coresets with a size only polylogarithmic in the number of data points $n$. Similarly, we obtain the following Theorem for Euclidean spaces.

**Theorem 4** (Euclidean one-shot coresets). *Let $k \in \mathbb{N}$, $\varepsilon \in (0,1)$, $\delta \in (0,1)$ and $p \in [1,p_{max}]$. Let $\mathcal{X}$ be a data set of $n$ points in $\mathbb{R}^d$, $\Delta = \frac{1}{\log n}$ and*

$$m \geq \frac{cS\log S}{\varepsilon^2}\left(dk\log k + \log\frac{1}{\delta\varepsilon} + \log\log n\right)$$

*where $c > 0$ is an absolute constant and $S = 16^{p_{max}}k\log p_{max}\log n$. Then, the weighted set $\mathcal{C}$ returned by Algorithm 4 is a $\varepsilon$-coreset of $\mathcal{X}$ for $k$-clustering in $\mathbb{R}^d$ for all $p \in [1, p_{max}]$, with probability at least $1 - \delta$. The computational complexity of Algorithm 4 is $\mathcal{O}(nkd\log n\log\frac{1}{\delta}\log p_{max})$.*

Theorem 4 implies one-shot coresets for Euclidean spaces with size near-logarithmic in the number of data points $n$, ignoring the $\log\log n$ factors. Furthermore, as in Theorem 2, the dependence on the number of clusters $k$ is of $\mathcal{O}(k^2\log^2 k)$ while the dependence on the ambient dimension $d$ is linear.

### 4.2 Analysis

The key technical difficulty in our proof of Theorems 3 and 4 is to show that Algorithm 4 computes coreset $\mathcal{C}$ uniformly for all $p \in [1,p_{max}]$.

We start by showing that the sensitivity for all $p \in [1, p_{max}]$ may be bounded by the sum of sensitivities for all grid points $P = \{1, (1+\Delta), (1+\Delta)^2, \ldots, p_{max}\}$. First, we consider a single interval in the grid, i.e. $[p_0, p_0(1+\Delta)]$ for some $p_0 \geq 0$. Lemma 2 shows that the sensitivity for any $p \in [p_0, p_0(1+\Delta)]$ may be bounded by the sum of sensitivities of $p_0$ and $p_0(1+\Delta)$ up to a constant factor of at most $n^\Delta$.

**Algorithm 4** ONE-SHOT CORESETS

**Require:** data set $\mathcal{X}$, $k$, $p_{max}$, $\Delta$, $m$, $\delta$
1: $\ell \leftarrow \left\lceil\frac{\log p_{max}}{\log(1+\Delta)}\right\rceil$
2: $P = \{1, (1+\Delta), (1+\Delta)^2, \ldots, (1+\Delta)^\ell\}$
3: $s = 0$
4: **for each** $p \in P$ **do**
5:    $s_p \leftarrow$ Run Algorithm 2 with $\delta/2\ell$
6:    $s(\cdot) \leftarrow s(\cdot) + n^\Delta s_p(\cdot)$
7: **for each** $x \in \mathcal{X}$ **do**
8:    $q(x) \leftarrow s(x)/\sum_{x'\in\mathcal{X}}s(x')$
9: $\mathcal{C} \leftarrow$ Sample $m$ weighted points from $\mathcal{X}$ where each point $x$ has weight $\frac{1}{|\mathcal{X}|mq(x)}$ and is sampled with probability $q(x)$.
10: **return** $\mathcal{C}$

**Lemma 2.** *Let $\mathcal{X}$ be a set of $n$ points and $p, \Delta \in \mathbb{R}_{\geq 0}$. Then, for any $\theta \in [0,1]$ and any $x \in \mathcal{X}$, it holds that*

$$\frac{d(x,Q)^{p(1+\theta\Delta)}}{\phi_\mathcal{X}^{p(1+\theta\Delta)}(Q)} \leq n^{\theta\Delta}\left((1-\theta)\frac{d(x,Q)^p}{\phi_\mathcal{X}^p(Q)} + \theta\frac{d(x,Q)^{p(1+\Delta)}}{\phi_\mathcal{X}^{p(1+\Delta)}(Q)}\right).$$

*Proof.* For $\theta \in \{0,1\}$, the claim trivially holds and we henceforth only consider $\theta \in (0,1)$. For positive reals $a,b,c,\alpha,\beta$, such that $\frac{1}{\alpha} + \frac{1}{\beta} = 1$, Young's inequality implies that

$$abc \leq c\left(\frac{a^\alpha}{\alpha} + \frac{b^\beta}{\beta}\right). \tag{18}$$

Consider the choice of $\alpha = \frac{1}{1-\theta} > 0$ and $\beta = \frac{1}{\theta} > 0$ for which it holds that $\frac{1}{\alpha} + \frac{1}{\beta} = 1 - \theta + \theta = 1$. Let

$$a = \left(\frac{d(x,Q)^p}{\phi_\mathcal{X}^p(Q)}\right)^{1-\theta}$$

$$b = \left(\frac{d(x,Q)^{p(1+\Delta)}}{\phi_\mathcal{X}^{p(1+\Delta)}(Q)}\right)^\theta$$

$$c = \frac{\phi_\mathcal{X}^p(Q)^{1-\theta}\phi_\mathcal{X}^{p(1+\Delta)}(Q)^\theta}{\phi_\mathcal{X}^{p(1+\theta\Delta)}(Q)}.$$

Then, by design, we have that

$$abc = \frac{d(x,Q)^{p(1+\theta\Delta)}}{\phi_\mathcal{X}^{p(1+\theta\Delta)}(Q)}.$$

as well as

$$\frac{a^\alpha}{\alpha} + \frac{b^\beta}{\beta} = (1-\theta)\frac{d(x,Q)^p}{\phi_\mathcal{X}^p(Q)} + \theta\frac{d(x,Q)^{p(1+\Delta)}}{\phi_\mathcal{X}^{p(1+\Delta)}(Q)}.$$

Hoelder's inequality implies that

$$\phi_{\mathcal{X}}^{p(1+\theta\Delta)}(Q) = \frac{1}{n} \sum_{x \in \mathcal{X}} d(x,C)^{p(1+\theta\Delta)}$$
$$\geq \left( \frac{1}{n} \sum_{x \in \mathcal{X}} d(x,C)^p \right)^{1+\theta\Delta} \quad (19)$$
$$= \phi_{\mathcal{X}}^p(Q)^{1+\theta\Delta}.$$

As a result,

$$c = \frac{\phi_{\mathcal{X}}^p(Q)^{1-\theta} \phi_{\mathcal{X}}^{p(1+\Delta)}(Q)^\theta}{\phi_{\mathcal{X}}^{p(1+\Delta)}(Q)} \leq \frac{\phi_{\mathcal{X}}^{p(1+\Delta)}(Q)^\theta}{\phi_{\mathcal{X}}^p(Q)^{(1+\Delta)\theta}}$$
$$= n^{\theta\Delta} \left( \frac{\sum_{x \in \mathcal{X}} d(x,C)^{p(1+\Delta)}}{\left( \sum_{x \in \mathcal{X}} d(x,C)^p \right)^{1+\Delta}} \right)^\theta \leq n^{\theta\Delta},$$

where the last inequality follows from

$$\frac{\sum_{x \in \mathcal{X}} d(x,C)^{p(1+\Delta)}}{\left( \sum_{x \in \mathcal{X}} d(x,C)^p \right)^{1+\Delta}} = \sum_{x \in \mathcal{X}} \underbrace{\left( \frac{d(x,C)^p}{\sum_{x' \in \mathcal{X}} d(x',C)^p} \right)^{1+\Delta}}_{\leq 1}$$

which concludes the proof. □

As Lemma 2 is critical for our proof, we show that it is tight up to constants.

**Lemma 3.** *For any $p > 0$, $\Delta \in (0,1]$ and $n \in \mathbb{N}$ sufficiently large, there exists a set $\mathcal{X}$ of $n$ points and a query $Q$ of at most $k$ points such that for some $x \in \mathcal{X}$*

$$\frac{d(x,Q)^{p(1+\Delta/2)}}{\phi_{\mathcal{X}}^{p(1+\Delta/2)}(Q)} \geq \frac{n^{\Delta/6}}{9} \left( \frac{d(x,Q)^p}{\phi_{\mathcal{X}}^p(Q)} + \frac{d(x,Q)^{p(1+\Delta)}}{\phi_{\mathcal{X}}^{p(1+\Delta)}(Q)} \right)$$

The proof is provided in the Appendix and is based upon an explicit construction of a data set $\mathcal{X}$ for which the claimed lower bound is achieved.

Lemma 4 uses Lemmas 1 and 2 to show that Algorithm 4 computes a valid sensitivity bound $s(x)$ for all $p \in [1, p_{max}]$.

**Lemma 4.** *Choose $\Delta = \frac{1}{\log n}$ and let $s : \mathcal{X} \to \mathbb{R}_{\geq 0}$ be computed as in Algorithm 4. Then, with probability at least $1 - \frac{\delta}{2}$, it holds that*

$$s(x) \geq \sup_{Q \in \mathcal{Q}, p \in [1, p_{max}]} \frac{d(x,Q)^p}{\frac{1}{n} \sum_{x' \in \mathcal{X}} d(x',Q)^p}, \quad (20)$$

*where $\mathcal{Q} = \mathbb{R}^{d \times k}$ for Euclidean spaces and $\mathcal{Q} = \mathcal{X}^k$ for general metric spaces. Furthermore, we have $|P| \in \mathcal{O}(\log n \log p_{max})$ and $S = \frac{1}{|\mathcal{X}|} \sum_{x \in \mathcal{X}} s(x) \in \mathcal{O}(8^{p_{max}} k \log n \log p_{max})$.*

*Proof.* Without loss of generality assume that $\ell = \frac{\log p_{max}}{\log(1+\Delta)}$ and thus $(1+\Delta)^\ell = p_{max}$. Using a union bound and applying Lemma 1 (with input probability equal to $\frac{\delta}{2\ell}$) for each $p \in P$, we obtain that with probability at least $1 - \frac{\delta}{2}$, each $s_p(\cdot) \geq \sigma_p(\cdot)$ and $S_p = \frac{1}{|\mathcal{X}|} \sum_{x \in \mathcal{X}} s_p(x) \in \mathcal{O}(8^p k)$. As a result, $s(x) = \sum_{p \in P} n^\Delta s_p(x)$ is a uniform upper bound on $\sigma_p(\cdot)$ for all $p \in P$. By Lemma 2 and since $\Delta = \frac{1}{\log n}$, $s(\cdot)$ is an upper bound on $\sigma_p(\cdot)$ for *all* $p \in [1, p_{max}]$. Since $\log(1+x) \geq \frac{x}{2}$, for $x \in [0,2]$, we have

$$\ell = \frac{\log p_{max}}{\log(1+\Delta)} \leq \frac{2 \log p_{max}}{\Delta} \in \mathcal{O}(\log n \log p_{max}).$$

Thus, we finally have

$$S = \sum_{p \in P} S_p \in \mathcal{O}(8^{p_{max}} k \log p_{max} \log n)$$

which concludes the proof.

□

Our last auxiliary result in Lemma 5 shows that if a set $\mathcal{C}$ is a coreset for all $p$ on a fine enough grid $\tilde{P}$, then it is a one-shot coreset for all $p \in [1, p_{max}]$.

**Lemma 5.** *Let $p_{max} > 1$ and $\varepsilon > 0$. Let $\mathcal{X}$ be a set of $n$ points. For $\gamma = \frac{\varepsilon}{6 \log n}$ and $r = \left\lfloor \frac{\log p_{max}}{\log(1+\gamma)} \right\rfloor \in \mathcal{O}\left(\frac{1}{\varepsilon} \log n \log p_{max}\right)$, define*

$$\tilde{P} = \left\{ 1, (1+\gamma), (1+\gamma)^2, \ldots, (1+\gamma)^r \right\}.$$

*If a weighted set $C$ is a $\frac{\varepsilon}{3}$-coreset of $\mathcal{X}$ for all $p \in \tilde{P} \cup \{p_{max}\}$, then it is also a $\varepsilon$-coreset of $\mathcal{X}$ for all $p \in [1, p_{max}]$.*

*Proof.* Without loss of generality, for the remainder of the proof we assume that $p_{max} = (1+\gamma)^r$. For any $p \in [1, p_{max}]$, there hence exists a $p' \in \tilde{P}$ and a $\theta \in [0,1]$ such that $p = p'(1+\theta\gamma)$. To prove the lemma, it is thus sufficient to show that for any query $Q$, any $\theta \in [0,1]$ and any $p' \in \tilde{P}$

$$\left| \phi_{\mathcal{X}}^{p'(1+\theta\gamma)}(Q) - \phi_{\mathcal{C}}^{p'(1+\theta\gamma)}(Q) \right| \leq \varepsilon \phi_{\mathcal{X}}^{p'(1+\theta\gamma)}(Q),$$

or equivalently

$$1 - \varepsilon \leq \frac{\phi_{\mathcal{C}}^{p'(1+\theta\gamma)}(Q)}{\phi_{\mathcal{X}}^{p'(1+\theta\gamma)}(Q)} \leq 1 + \varepsilon. \quad (21)$$

Using Lemma 2 and summing both sides across the weighted set $\mathcal{C}$, we obtain

$$\frac{\phi_\mathcal{C}^{p'(1+\theta\gamma)}(Q)}{\phi_\mathcal{X}^{p'(1+\theta\gamma)}(Q)} \leq n^{\theta\gamma}\left((1-\theta)\frac{\phi_\mathcal{C}^{p'}(Q)}{\phi_\mathcal{X}^{p'}(Q)} + \theta\frac{\phi_\mathcal{C}^{p'(1+\gamma)}(Q)}{\phi_\mathcal{X}^{p'(1+\gamma)}(Q)}\right).$$

Since $\mathcal{C}$ is a $\frac{\varepsilon}{3}$-coreset for both $p'$ and $p'(1+\gamma)$, both fractions on the right-hand side are bounded by $1+\frac{\varepsilon}{3}$. Since $\log(1+x) \geq \frac{x}{2}$ for $x \in [0,1]$, or equivalently $e^{\frac{x}{2}} \leq 1+x$, we have that

$$n^{\theta\gamma} = e^{\frac{\theta\varepsilon}{6}} \leq e^{\frac{\varepsilon}{6}} \leq 1 + \frac{\varepsilon}{3}.$$

We conclude that

$$\frac{\phi_\mathcal{C}^{p'(1+\theta\gamma)}(Q)}{\phi_\mathcal{X}^{p'(1+\theta\gamma)}(Q)} \leq \left(1+\frac{\varepsilon}{3}\right)^2 \stackrel{(\varepsilon<1)}{\leq} 1 + \varepsilon$$

which proves the upper bound in (21).

For any $x \in \mathcal{X}$, we have

$$\frac{\mathrm{d}(x,Q)^{p'(1+\theta\gamma)}}{n\phi_\mathcal{X}^{p'(1+\theta\gamma)}(Q)} \geq \frac{\mathrm{d}(x,Q)^{p'(1+\gamma)}}{\left[n\phi_\mathcal{X}^{p'(1+\theta\gamma)}(Q)\right]^{\frac{1+\gamma}{1+\theta\gamma}}}$$

since the term on the left is smaller or equal to one and $1+\gamma \geq 1+\theta\gamma$. Analogous to (19), Hoelder's inequality implies

$$\left[\phi_\mathcal{X}^{p'(1+\theta\gamma)}(Q)\right]^{\frac{1+\gamma}{1+\theta\gamma}} \leq \phi_\mathcal{X}^{p'(1+\gamma)}(Q)$$

while $\frac{1+\gamma}{1+\theta\gamma} \leq 1+\gamma$ since $\theta \geq 0$. As a result, we have for all $x \in \mathcal{X}$

$$\frac{\mathrm{d}(x,Q)^{p'(1+\theta\gamma)}}{\phi_\mathcal{X}^{p'(1+\theta\gamma)}(Q)} \geq n^{-\gamma}\frac{\mathrm{d}(x,Q)^{p'(1+\gamma)}}{\phi_\mathcal{X}^{p'(1+\gamma)}(Q)}$$

Summing both sides across the weighted set $\mathcal{C}$ and using that $\mathcal{C}$ is a $\frac{\varepsilon}{3}$-coreset for $p'(1+\gamma)$ as well as $n^{-\gamma} \geq 1 - \frac{\varepsilon}{3}$, we obtain

$$\frac{\phi_\mathcal{C}^{p'(1+\theta\gamma)}(Q)}{\phi_\mathcal{X}^{p'(1+\theta\gamma)}(Q)} \geq \left(1-\frac{\varepsilon}{3}\right)\frac{\phi_\mathcal{C}^{p'(1+\gamma)}(Q)}{\phi_\mathcal{X}^{p'(1+\gamma)}(Q)} \geq 1 - \varepsilon$$

which shows the lower bound in (21). $\square$

Lemmas 4 and 5 are sufficient to prove both Theorems 3 and 4.

*Proof of Theorem 3.* Let $\tilde{P}$ be defined as in Lemma 5 and note that $|\tilde{P}| \in \mathcal{O}\left(\frac{1}{\varepsilon}\log n \log p_{max}\right)$.

By Lemma 4, $s(\cdot)$ as computed in Algorithm 4 provides a uniform upper bound on the sensitivity $\sigma_p(\cdot)$ for all $p \in [1, p_{max}]$ and hence also $p \in \tilde{P}$. Furthermore, we have that $S \in \mathcal{O}(8^{p_{max}} k \log n \log p_{max})$.

For a single $p \in \tilde{P}$, we now follow the proof of Theorem 1 but with $\varepsilon' = \frac{\varepsilon}{3}$, $\delta' = \frac{\delta}{|\tilde{P}|}$ and the sensitivity bound $s(\cdot)$ instead of $s_p(\cdot)$. For

$$m \in \Omega\left(\frac{S}{\varepsilon'^2}\left(k\log n + \log\frac{1}{\delta'}\right)\right)$$

$\mathcal{C}$ is a $\frac{\varepsilon}{3}$-coreset of $\mathcal{X}$ for a single $p \in \tilde{P}$ with probability at least $1 - \frac{\delta}{|\tilde{P}|}$. By the union bound, with probability at least $1-\delta$, $\mathcal{C}$ is a $\frac{\varepsilon}{3}$-coreset of $\mathcal{X}$ for *all* $p \in \tilde{P}$. By Lemma 5 this implies that $\mathcal{C}$ is a $\varepsilon$-coreset of $\mathcal{X}$ for $p \in [1, p_{max}]$ as claimed. The computational complexity of Algorithm 4 is dominated by $\ell \in \mathcal{O}(\log n \log p_{max})$ calls to Algorithm 3 where each invocation has a complexity of $\mathcal{O}\left(nkd\log\frac{1}{\delta}\right)$. $\square$

*Proof of Theorem 4.* The proof follows directly from the proof of Theorem 3 with the following variation: To show that $\mathcal{C}$ is a $\frac{\varepsilon}{3}$-coreset of $\mathcal{X}$ for each $p \in \tilde{P}$, we use the proof of Theorem 2 and not of Theorem 1. $\square$

## 5 Conclusion

We have presented a state-of-the-art coreset construction for a large family of $k$-clustering problems in both Euclidean and general metric spaces. Specific instances of that family include $k$-Median and $k$-Means clustering. We have further introduced the notion of one-shot coresets which are data summaries suitable for an unbounded number of $k$-clustering problems at once. We have shown how to construct such one-shot coresets for both Euclidean and general metric spaces and have provided bounds on the required coreset size.

### Acknowledgments

This research was partially supported by SNSF NRP 75 (project number 407540_167212), ERC StG 307036, and a Google Ph.D. Fellowship. This work was done in part while Olivier Bachem was at Google Research New York.

# References

[1] Pankaj K Agarwal, Sariel Har-Peled, and Kasturi R Varadarajan. Geometric approximation via coresets. *Combinatorial and computational geometry*, 52:1–30, 2005.

[2] David Arthur and Sergei Vassilvitskii. k-means++: The advantages of careful seeding. In *Symposium on Discrete Algorithms (SODA)*, pages 1027–1035. SIAM, 2007.

[3] Olivier Bachem, Mario Lucic, and Andreas Krause. Coresets for nonparametric estimation - the case of DP-means. In *International Conference on Machine Learning (ICML)*, 2015.

[4] Olivier Bachem, Mario Lucic, S. Hamed Hassani, and Andreas Krause. Approximate k-means++ in sublinear time. In *Conference on Artificial Intelligence (AAAI)*, 2016.

[5] Olivier Bachem, Mario Lucic, S. Hamed Hassani, and Andreas Krause. Fast and provably good seedings for k-means. In *Neural Information Processing Systems (NIPS)*, 2016.

[6] Olivier Bachem, Mario Lucic, S Hamed Hassani, and Andreas Krause. Uniform deviation bounds for k-means clustering. In *International Conference on Machine Learning (ICML)*, pages 283–291, 2017.

[7] Olivier Bachem, Mario Lucic, and Andreas Krause. Scalable and distributed clustering via lightweight coresets. *arXiv preprint arXiv:1702.08248*, 2017.

[8] Olivier Bachem, Mario Lucic, and Andreas Krause. Practical coreset constructions for machine learning. *arXiv preprint arXiv:1703.06476*, 2017.

[9] Bahman Bahmani, Benjamin Moseley, Andrea Vattani, Ravi Kumar, and Sergei Vassilvitskii. Scalable k-means++. *International Conference on Very Large Data Bases (VLDB)*, 5(7):622–633, 2012.

[10] Maria-Florina F Balcan, Steven Ehrlich, and Yingyu Liang. Distributed $k$-means and $k$-median clustering on general topologies. In *Neural Information Processing Systems (NIPS)*, pages 1995–2003, 2013.

[11] Vladimir Braverman, Dan Feldman, and Harry Lang. New frameworks for offline and streaming coreset constructions. *arXiv preprint arXiv:1612.00889*, 2016.

[12] Ke Chen. On coresets for k-median and k-means clustering in metric and euclidean spaces and their applications. *SIAM Journal on Computing*, 39(3):923–947, 2009.

[13] Sanjoy Dasgupta. *The hardness of k-means clustering*. Department of Computer Science and Engineering, University of California, San Diego, 2008.

[14] Dan Feldman, Melanie Schmidt, and Christian Sohler. Turning big data into tiny data: Constant-size coresets for k-means, pca and projective clustering. In *Symposium on Discrete Algorithms (SODA)*, pages 1434–1453. SIAM, 2013.

[15] Michael Langberg and Leonard J Schulman. Universal $\varepsilon$-approximators for integrals. In *Symposium on Discrete Algorithms (SODA)*, pages 598–607. SIAM, 2010.

[16] Yi Li, Philip M Long, and Aravind Srinivasan. Improved bounds on the sample complexity of learning. *Journal of Computer and System Sciences*, 62(3):516–527, 2001.

[17] Mario Lucic, Olivier Bachem, and Andreas Krause. Strong coresets for hard and soft Bregman clustering with applications to exponential family mixtures. In *International Conference on Artificial Intelligence and Statistics (AISTATS)*, pages 1–9, 2016.

[18] Mario Lucic, Matthew Faulkner, Andreas Krause, and Dan Feldman. Training mixture models at scale via coresets. *arXiv preprint arXiv:1703.08110*, 2017.

[19] David Pollard. A few good inequalities. Technical report, 2015.

## A  Proof of Lemma 1

*Proof.* By Theorem 5.5 of Arthur and Vassilvitskii [2], the solution $C$ returned by a single run of Algorithm 1 satisfies

$$\mathbb{E}[\phi^p_{\mathcal{X}}(C)] \leq 2^{p+2} \left(\log_2 k + 2\right) \phi^p_{\mathcal{X}}(B^*)$$

where $B^*$ denotes the optimal solution. Let $B$ be the best of $\ln \frac{1}{\delta}$ runs of Algorithm 1. Then, Markov's inequality implies that with probability at most $2^{-\ln \frac{1}{\delta}} \leq \delta$

$$\phi^p_{\mathcal{X}}(B) > 2^{p+3} \left(\log_2 k + 2\right) \phi^p_{\mathcal{X}}(B^*).$$

Hence, with probability at least $1 - \delta$, $B$ is $\alpha \leq 2^{p+3} \left(\log_2 k + 2\right)$ competitive compared to the optimal solution $B^*$.

Fix $x \in \mathcal{X}$ and $Q \in \mathbb{R}^{d \times k}$ for Euclidean spaces or $Q \in \mathcal{X}^k$ for general metric spaces. Let $b_x$ be the center in $B$ that is closest to $x$ with ties broken arbitrarily but consistently. Similarly, let $X_x$ be all the points $x \in X$ for which $b_x$ is the closest cluster center. The generalized triangle inequality implies

$$d(b_x, Q)^p \leq 2^{p-1} \left[d(x, b_x)^p + d(x, Q)^p\right].$$

Summing over all $x' \in X_x$ and dividing by $\frac{1}{|X_x|}$, we obtain that

$$d(b_x, Q)^p \leq 2^{p-1} \phi^p_{\mathcal{X}_x}(b_x) + 2^{p-1} \phi^p_{\mathcal{X}_x}(Q).$$

Together with the generalized triangle inequality, this yields

$$\frac{d(x, Q)^p}{2^{2p-2}} \leq 2^{1-p}[d(x, b_x)^p + d(b_x, Q)^p]$$
$$\leq 2^{1-p} d(x, b_x)^p + \phi^p_{\mathcal{X}_x}(b_x) + \phi^p_{\mathcal{X}_x}(Q).$$

Dividing both sides by $\phi^p_{\mathcal{X}}(Q)$, we obtain

$$\frac{d(x,Q)^p}{4^{p-1}\phi^p_{\mathcal{X}}(Q)} \leq \frac{2^{1-p}d(x,b_x)^p}{\phi^p_{\mathcal{X}}(Q)} + \frac{\phi^p_{\mathcal{X}_x}(b_x)}{\phi^p_{\mathcal{X}}(Q)} + \frac{\phi^p_{\mathcal{X}_x}(Q)}{\phi^p_{\mathcal{X}}(Q)}$$
$$\leq \frac{\alpha 2^{1-p}d(x,b_x)^p}{\phi^p_{\mathcal{X}}(B)} + \alpha \frac{\phi^p_{\mathcal{X}_x}(b_x)}{\phi^p_{\mathcal{X}}(B)} + \frac{|\mathcal{X}|}{|\mathcal{X}_x|}.$$

where the last inequality follows since $B$ is $\alpha$ optimal with regards to the optimal solution and by the definitions of $\phi^p_{\mathcal{X}_x}(Q)$ and $\phi^p_{\mathcal{X}}(Q)$. This proves that $s_p(\cdot)$ as returned by Algorithm 2 is a valid upper bound for $\sigma_p(x)$ as claimed.

Furthermore, summing $s_p(x)$ across $x \in \mathcal{X}$ leads to

$$\frac{1}{|\mathcal{X}|} \sum_{x \in \mathcal{X}} s_p(x) = \alpha 2^{p-1} + \alpha 2^{2p-2} + 2^{2p-2} k$$

Since $\log_2 k + 2 \leq 2k$, we have $\alpha \leq 2^{p+4} k$ and hence

$$\frac{1}{|\mathcal{X}|} \sum_{x \in \mathcal{X}} s_p(x) \leq 2^{2p+3} k + 2^{3p+2} k + 2^{2p-2} k$$

which shows the claim of $S \leq 8^{p+2} k$. Finally, the computational complexity of Algorithm 2 is dominated by running Algorithm 1 $\mathcal{O}\left(\log \frac{1}{\delta}\right)$ times where each run has a complexity of $\mathcal{O}(ndk)$. □

## B  Proof of Theorem 2

Our proof of Theorem 2 relies on the notion of *pseudo-dimension* — a generalization of the *Vapnik-Chervonenkis dimension* to $[0, 1]$-valued functions — and the following seminal result by Li et al. [16].

**Definition 5.** *[Pseudo-dimension] Fix a countably infinite domain $\mathcal{X}$. The pseudo-dimension of a set $\mathcal{F}$ of functions from $\mathcal{X}$ to $[0,1]$, $\operatorname{Pdim}(\mathcal{F})$, is the largest $d'$ such there is a sequence $x_1, \ldots, x_{d'}$ of domain elements from $\mathcal{X}$ and a sequence of reals $r_1, \ldots, r_{d'}$ of real thresholds such that for each $b_1, \ldots, b_{d'} \in \{above, below\}$, there is an $f \in \mathcal{F}$ such that for all $i = 1, \ldots, d'$, we have $f(x_i) \geq r_i \iff b_i = above$.*

**Theorem 5** (Li et al. [16])**.** *Let $\alpha > 0$, $\nu > 0$ and $\delta > 0$. Fix a countably infinite domain $\mathcal{X}$ and let $q(\cdot)$ be any probability distribution over $\mathcal{X}$. Let $\mathcal{F}$ be a set of functions from $\mathcal{X}$ to $[0, 1]$ with $\operatorname{Pdim}(\mathcal{F}) = d'$. Denote by $\mathcal{C}$ a sample of $m$ points from $\mathcal{X}$ sampled independently according to $q(\cdot)$. Then, for $m \geq \frac{c}{\alpha^2 \nu}(d' \log \frac{1}{\nu} + \log \frac{1}{\delta})$ where $c$ is an absolute constant, it holds with probability at least $1 - \delta$ that*

$$\forall f \in \mathcal{F}: \quad \mathrm{d}_\nu \left( \sum_{x \in \mathcal{X}} q(x) f(x), \frac{1}{|\mathcal{C}|} \sum_{x \in \mathcal{C}} f(x) \right) \leq \alpha$$

*where $\mathrm{d}_\nu(a, b) = \frac{|a - b|}{a + b + \nu}$.*

*Proof of Theorem 2.* Our proof of Theorem 2 follows closely the proof of Theorem 1. However, instead of showing (9) using the Bernstein inequality and a union bound, we will use Theorem 5.

Let $g^p_Q(x)$ be as defined in (6) and define the family of $[0,1]$-bounded functions $\mathcal{F} =$

$\left\{g_Q^p(\cdot) \mid Q \in \mathbb{R}^{d \times k}\right\}$. An adaptation[1] of the proof of Lemma 1 in Bachem et al. [6] yields that $\text{Pdim}(\mathcal{F}) \in \mathcal{O}(dk \log k)$.

Choose $\nu = \frac{1}{4S}$, $\alpha = \frac{\varepsilon}{4}$ and $q(x)$ defined as in Theorem 1. We apply Theorem 5 to the function family $\mathcal{F}$. This implies that for $m \geq \frac{cS}{\varepsilon^2} \log S \left(dk \log k + \log \frac{1}{\delta}\right)$, we have with probability at least $1 - \frac{\delta}{2}$, for all $Q \in \mathbb{R}^{d \times k}$

$$d_{\frac{1}{4S}}\left(\sum_{x \in \mathcal{X}} q(x) g_Q^p(x), \frac{1}{|\mathcal{C}|} \sum_{x \in \mathcal{C}} g_Q^p(x)\right) \leq \frac{\varepsilon}{4}.$$

Note that $\sum_{x \in \mathcal{X}} p(x) g_Q^p(x) = \frac{1}{S}$ and define $\beta = \frac{S}{|\mathcal{C}|} \sum_{x \in \mathcal{C}} g_Q^p(x)$. This then implies using the definition of $d_\nu(\cdot, \cdot)$ that

$$\left|\frac{1}{S} - \frac{\beta}{S}\right| \leq \left(\frac{1}{S} + \frac{\beta}{S} + \frac{1}{4S}\right)\frac{\varepsilon}{4}$$

or equivalently

$$|1 - \beta| \leq \frac{5}{16}\varepsilon + \frac{\varepsilon}{4}\beta. \qquad (22)$$

On one hand, this implies that

$$1 - \beta \leq \frac{5}{16}\varepsilon + \frac{\varepsilon}{4}\beta \Leftrightarrow -\frac{4+\varepsilon}{4}\beta \leq \frac{5}{16}\varepsilon - 1$$

and hence

$$-\beta \leq \frac{\frac{5}{4}\varepsilon - 4}{4 + \varepsilon} \Leftrightarrow 1 - \beta \leq \frac{\frac{9}{4}\varepsilon}{4 + \varepsilon} \leq \epsilon.$$

On the other hand, (22) implies that

$$\beta - 1 \leq \frac{5}{16}\varepsilon + \frac{\varepsilon}{4}\beta \Leftrightarrow \frac{4-\varepsilon}{4}\beta \leq \frac{5}{16}\varepsilon + 1$$

and thus

$$\beta \leq \frac{\frac{5}{4}\varepsilon + 4}{4 - \varepsilon} \Leftrightarrow \beta - 1 \leq \frac{\frac{9}{4}\varepsilon}{4 - \varepsilon} \leq \varepsilon$$

where the last inequality follows since $\varepsilon \leq 1$. Combining these two results, we obtain that with probability at least $1 - \frac{\delta}{2}$, $|1 - \beta| \leq \varepsilon$. By definition of $\beta$ and using that $S \in \mathcal{O}(8^p k)$ by Lemma 1 and $\frac{\phi_{\mathcal{C}}^p(Q)}{S\phi_{\mathcal{X}}^p(Q)} = \frac{1}{|\mathcal{C}|} \sum_{x \in \mathcal{C}} g_Q^p(x)$, we then have that (9) holds with probability at least $1 - \frac{\delta}{2}$ which concludes the proof. □

---

[1] We simply replace the function $f_Q(x)$ in Bachem et al. [6] with the function $g_Q^p(x)$ as in (6) and note that for fixed $p$ the dichotomies induced by a set of centers $Q$ is independent of $p$.

## C  Proof of Lemma 3

*Proof.* To prove the claim, we will construct a data set in one-dimensional Euclidean space for which Lemma 2 is tight up to constant factors. Let $x_1 = n^{\frac{1}{p(2+\Delta)}}$, $x_2 = 1$ and $x_3 = n^{-\frac{1}{p(2+\Delta)}}$. Define the query $Q = 0$ and note that $d(x_1, Q)^p = n^{\frac{1}{2+\Delta}}$, $d(x_2, Q)^p = 1$ and $d(x_3, Q)^p = n^{-\frac{1}{2+\Delta}}$. Let the data set $\mathcal{X}$ consist of one copy of $x_1$, $\sqrt{n}$ copies of $x_2$ and $\beta = n - \sqrt{n} - 1$ copies of $x_3$. Now, $\mathcal{X}$ consists of $n$ points as claimed. For $n$ sufficiently large, it holds that $\beta = n - \sqrt{n} - 1 \geq \frac{1}{2}n$ and hence

$$n\phi_{\mathcal{X}}^p(Q) = n^{\frac{1}{2+\Delta}} + \sqrt{n} + \beta n^{-\frac{1}{2+\Delta}}$$
$$\geq \frac{1}{2} n^{\frac{1+\Delta}{2+\Delta}}. \qquad (23)$$

Similarly, we have that

$$n\phi_{\mathcal{X}}^{p(1+\Delta)}(Q) = n^{\frac{1+\Delta}{2+\Delta}} + \sqrt{n} + \beta n^{-\frac{1+\Delta}{2+\Delta}}$$
$$\geq n^{\frac{1+\Delta}{2+\Delta}}. \qquad (24)$$

On the other hand, it holds that

$$n\phi_{\mathcal{X}}^{p(1+\Delta/2)}(Q) = n^{\frac{1+\Delta/2}{2+\Delta}} + \sqrt{n} + \beta n^{-\frac{1+\Delta/2}{2+\Delta}}$$
$$\leq 3\sqrt{n}. \qquad (25)$$

Using (23) and (24), we have that the right-hand side in Lemma 3 is bounded from above by

$$\texttt{RHS} = \frac{n^{\Delta/6}}{9}\left(\frac{d(x_2, Q)^p}{\phi_{\mathcal{X}}^p(Q)} + \frac{d(x_2, Q)^{p(1+\Delta)}}{\phi_{\mathcal{X}}^{p(1+\Delta)}(Q)}\right)$$
$$= \frac{n^{\Delta/6}}{9}\left(\frac{1}{\phi_{\mathcal{X}}^p(Q)} + \frac{1}{\phi_{\mathcal{X}}^{p(1+\Delta)}(Q)}\right) \qquad (26)$$
$$\leq \frac{1}{3} n^{\Delta/6 + 1 - \frac{1+\Delta}{2+\Delta}}.$$

As $0 \leq \Delta \leq 1$, it follows that

$$\frac{\Delta}{6} + 1 - \frac{1+\Delta}{2+\Delta} = \underbrace{\frac{\Delta}{6} - \frac{\Delta}{2(2+\Delta)}}_{\leq 0} + \frac{1}{2} \leq \frac{1}{2}.$$

Together with (25), this implies that

$$\frac{1}{3} n^{\Delta/6 + 1 - \frac{1+\Delta}{2+\Delta}} \leq \frac{1}{3}\sqrt{n} \leq \frac{1}{\phi_{\mathcal{X}}^{p(1+\Delta/2)}(Q)}$$
$$\leq \frac{d(x_2, Q)^{p(1+\Delta/2)}}{\phi_{\mathcal{X}}^{p(1+\Delta/2)}(Q)} = \texttt{LHS}. \qquad (27)$$

Hence, combining (26) with (27) implies that the main claim $\texttt{LHS} \geq \texttt{RHS}$ holds for $x = x_2$. □